\documentclass[review]{elsarticle}

\usepackage{hyperref}
\usepackage{times}
\usepackage{epsfig}
\usepackage{amsmath}
\usepackage{amssymb}
\usepackage{mathtools}
\usepackage{multirow}
\usepackage[table,xcdraw, dvipsnames]{xcolor}
\usepackage{xcolor}
\usepackage{multirow}
\usepackage[caption=false]{subfig}

\journal{}
\biboptions{authoryear}

\begin{document}
\begin{frontmatter}
\title{Exploiting Inter-Frame Regional Correlation\\ for Efficient Action Recognition}
\author[1]{Yuecong Xu\corref{mycorrespondingauthor}}
\cortext[mycorrespondingauthor]{Corresponding author}
\ead{xuyu0014@e.ntu.edu.sg}
\author[1]{Jianfei Yang\corref{}}
\ead{yang0478@e.ntu.edu.sg}
\author[1]{Kezhi Mao\corref{}}
\ead{ekzmao@ntu.edu.sg}
\author[2]{Jianxiong Yin\corref{}}
\ead{jianxiongy@nvidia.com}
\author[2]{Simon See\corref{}}
\ead{ssee@nvidia.com}
\address[1]{School of Electrical and Electronic Engineering, Nanyang Technological University,\\ 50 Nanyang Avenue, 639798, Singapore.}
\address[2]{NVIDIA AI Tech Centre,\\ 3 International Business Park Rd, \#01-20A Nordic European Centre, 609927, Singapore.}

\begin{abstract}
    Temporal feature extraction is an important issue in video-based action recognition. Optical flow is a popular method to extract temporal feature, which produces excellent performance thanks to its capacity of capturing pixel-level correlation information between consecutive frames. However, such a pixel-level correlation is extracted at the cost of high computational complexity and large storage resource. In this paper, we propose a novel temporal feature extraction method, named Attentive Correlated Temporal Feature (ACTF), by exploring inter-frame correlation within a certain region. The proposed ACTF exploits both bilinear and linear correlation between successive frames on the regional level. Our method has the advantage of achieving performance comparable to or better than optical flow-based methods while avoiding the introduction of optical flow. Experimental results demonstrate our proposed method achieves the state-of-the-art performances of $96.3\%$ on UCF101 and $76.3\%$ on HMDB51 benchmark datasets.
\end{abstract}


\end{frontmatter}


\section{Introduction}

Action recognition has received considerable attention from the vision community in recent years~\citep{HERATH20174, yang2019asymmetric, carmona2018human, wang2018learning, KECELI2018235, SAHOO2019524} thanks to its increasing applications in various fields, such as surveillance~\citep{danafar2007action, yang2018carefi, xiang2008activity, li2017accurate, BENMABROUK2018480, KARDAS2017343} and smart homes~\citep{wu2010multiview, 8391737, ortis2017organizing, LUNDSTROM2016429, LEE2017299} etc. Compared with static images, videos contain additional temporal information. Hence, extracting and handling temporal information is very critical in action recognition.
 
To extract temporal features underlying a video, a few methods have been proposed in the literature. Most of these efforts can be organized into two categories. The first category is the two-stream methods. A typical work in this category is the one proposed in~\citep{NIPS2014_5353}, which conduct the classification using temporal features and spatial features separately. The two types of features are integrated through classification decision fusion. The second category is the 3D ConvNet methods, which extract spatial and temporal features jointly by expanding the convolution kernel of 2D ConvNets to the temporal dimension. A seminal work in this category is the C3D~\citep{Tran:2015:LSF:2919332.2919929} network. A detailed review of the methods in both categories is given in Section~\ref{section:related}.
 
The methods in the two categories have their respective merits and limitations. The two-stream methods often produce the state-of-the-art performance, yet this is achieved at the cost of heavy reliance on accurate temporal feature. Therefore, the two-stream methods usually involve computation or estimation of optical flow, both of which require high computational power and large storage resource.  Also, obtaining optical flow needs to be performed prior to the training of the network, thus methods utilizing optical flow cannot be trained end-to-end. On the other hand, the 3D ConvNets-based methods are computationally less demanding, yet their performances are usually inferior to that of the two-stream methods. A possible reason would be the temporal pooling used for dimension reduction towards the complete representation. Temporal pooling extracts only linear feature along the temporal dimension of the video through pooling operation. With only the linear feature being extracted, we argue that part of the temporal feature is lost during the pooling operation.

\begin{figure}
    \centering
    \subfloat[Action "Handstand"\label{intuition-1}]{
    \includegraphics[width=1.\textwidth]{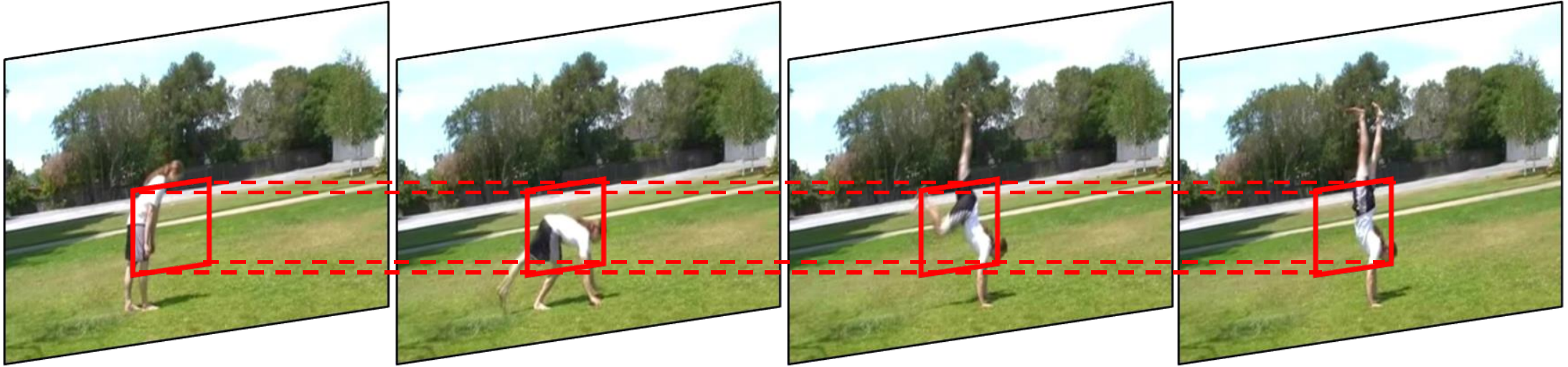}
    }
    \\
    \subfloat[Action "Brushing Teeth"\label{intuition-2}]{
    \includegraphics[width=1.\textwidth]{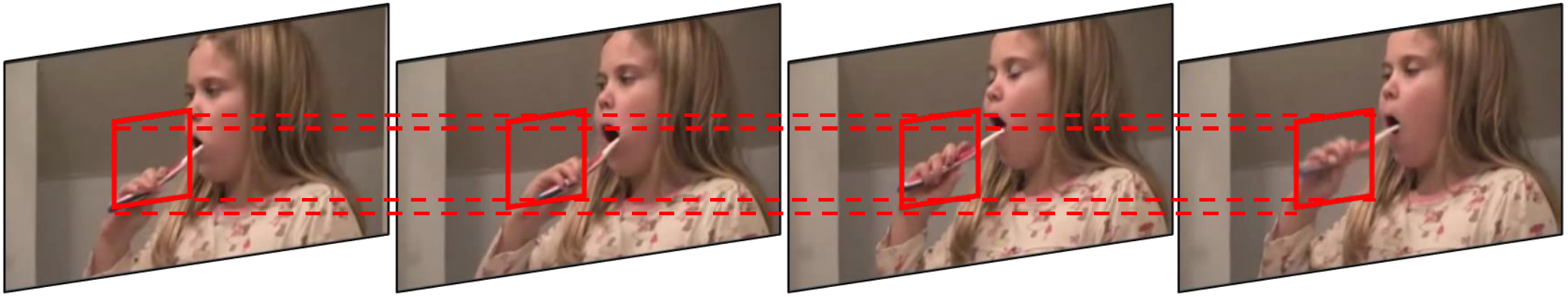}
    }
    \\
    \caption{Illustration of extracting inter-frame corresponding-regional correlation for action recognition. The temporal feature of an action is related to the correlation appearance between frames. Actions that are faster such as "Handstand" in (a) exhibits obvious change within the indicated box. Slower and more static actions such as "Brushing Teeth" in (b) shows little change between frames. To cope with both situations, bilinear operation is employed to extract the inter-frame corresponding-regional correlation}
    \label{introduction:intuition}
\end{figure}

In this paper, we present a novel method for temporal feature extraction, which achieves performance comparable to or even better than two-stream methods, yet demands less computational power. Intuitively, temporal feature of an action is related to the correlation of appearance between frames within a certain region. For instance, in Figure~\ref{intuition-1}, the indicated box across the series of frames shows how a person turns upside down, and is related to the action of "Handstand". Therefore, instead of using optical flow, our proposed method extracts temporal features by extracting the correlation of neighbouring frames with respect to the corresponding regions. The degree of change of appearance varies between different actions. For actions that are slower or more static, neighboring sampled frames could be very similar. One example is the action of "Brushing Teeth" shown in Figure~\ref{intuition-2}. If linear correlation, such as the difference in RGB value is employed, the correlation extracted would fail to contain temporal information of the video. To cope with the various type of actions, the inter-frame correlation would thus be computed through bilinear operation. The complete temporal feature, named as Attentive Correlated Temporal Feature (ACTF), is obtained through attentive combination of the inter-frame corresponding-regional correlation feature and the inter-frame mean feature obtained through inter-frame temporal average pooling.
Our main contributions are summarized as follows:
\renewcommand{\labelitemi}{$\ast$}
\begin{itemize}
    \item We propose a novel temporal feature extraction method: Attentive Correlated Temporal Feature (ACTF), for action recognition. First, ACTF exploits inter-frame corresponding-regional correlation to implicitly capture temporal information without the use of optical flow. Second, by excluding optical flow estimation or calculation, ACTF can be combined with any spatial feature extraction network under the two-stream structure to implement end-to-end training. Third, ACTF leads to performance comparable to or even better than optical flow-based methods, yet it demands less computation and memory due to the exclusion of optical flow.
    \item We conduct extensive experiments on two action recognition benchmark datasets: UCF101~\citep{soomro2012ucf101} and HMDB51~\citep{kuehne2011hmdb} with a framework utilizing our proposed ACTF. The results demonstrate that our proposed ACTF brings noticeable improvements over baseline methods, achieving state-of-the-art performance for these datasets.
\end{itemize}
The rest of this paper is organized as follows. Related works for action recognition tasks and the use of regional correlation are discussed in Section~\ref{section:related}. In Section~\ref{section:method}, we introduce the proposed Attentive Correlated Temporal Feature (ACTF) in detail. After that, we present and analyze the experimental results of our proposed ACTF feature, with a thorough ablation study on the design of ACTF. Finally, we conclude the paper in Section~\ref{section:conclusion}.

\section{Related Work}
\label{section:related}

Action recognition is one of the core tasks in video understanding. Compared to image understanding tasks, video understanding tasks are more complex due to the additional temporal dimension in videos. The extraction and handling of temporal feature underlying videos is thus the main challenge of the action recognition task.

\subsection{Action Recognition with Optical Flow}\label{related:optical-flow}
To extract temporal feature with high quality, previous works~\citep{NIPS2014_5353,NIPS2016_6433,wang2016temporal} adopt a two-stream strategy where temporal feature is extracted in parallel with spatial feature. The temporal feature extraction is performed by feeding a stack of optical flow frames computed by TV-L1~\citep{10.1007/978-3-540-74936-3_22} to a ConvNet. TSN~\citep{wang2016temporal} improves performance of the original two-stream network~\citep{NIPS2014_5353} through segmenting the video and input the RGB alongside optical flow frames for each segment to ConvNets with shared parameter. The action recognition result is produced through segmental consensus fusion. Meanwhile, ST-ResNet~\citep{NIPS2016_6433} builds upon ResNet~\citep{7780459}, which is a deeper 2D ConvNet. These two-stream methods achieve competitive results in action recognition, but using optical flow for temporal feature extraction has limitations. Optical flow extraction is known to be computationally expensive and memory intensive. In addition, as optical flow requires pre-computation, the use of optical flow as temporal feature prohibits fully end-to-end training of the network. 

\subsection{Temporal Feature Extraction without Optical Flow}\label{related:non-optical-flow}
To address the limitations imposed by utilizing optical flow, subsequent works proposed to extract temporal feature to replace optical flow. One category of methods involves the estimation of optical flow through neural network. FlowNet~\citep{7410673,8099662}, MotionNet~\citep{zhao2018through}, LMoF~\citep{li2018learn}, TVNet~\citep{fan2018end} and more recently Representation Flow~\citep{piergiovanni2019representation} all belong to such category. More specifically, FlowNet~\citep{7410673} learns optical flow from synthetic ground truth data. MotionNet~\citep{zhao2018through} produces optical flow through next frame prediction. LMoF~\citep{li2018learn} further constructs a learnable directional filtering layer to cope with optical flow estimation in blur videos. To further boost the performance of optical flow estimation, TVNet~\citep{fan2018end} unfolds the TV-L1~\citep{10.1007/978-3-540-74936-3_22} optical flow extraction method and formulates it with neural network. Representation Flow~\citep{piergiovanni2019representation} extends from TVNet~\citep{fan2018end} and constructs fully-differentiable convolutional layers to estimate optical flow. The layers could be stacked on top of each other to obtain flow-of-flow features which could capture longer-term motion representation. Although these optical flow estimation methods render networks to be trained in an end-to-end manner, they are still expensive in computation and intensive in memory, with longer run-time during inference. 
Another category extracts temporal feature jointly with spatial feature by constructing 3D ConvNets. C3D~\citep{Tran:2015:LSF:2919332.2919929}, 3D-ResNet\citep{tran2018closer}, 3D-ResNext~\citep{hara2018can}, I3D~\citep{carreira2017quo} and Asymmetric 3D-CNN~\citep{yang2019asymmetric} belong to this category. More specifically, C3D~\citep{Tran:2015:LSF:2919332.2919929} is one of the primary works where the CNN network is expanded to the temporal dimension. Subsequent networks such as 3D-ResNet~\citep{tran2018closer} and 3D-ResNext~\citep{hara2018can} are deeper and larger 3D ConvNets. To further reduce parameter for even faster training, Carreira \textit{et at.} inflates 2D ConvNets into 3D structure. This simplifies the work of constructing 3D ConvNets by simply convert image classification models to 3D models by endowing filters and pooling kernels with the additional temporal dimension. Whereas Yang \textit{et al.}~\citep{yang2019asymmetric} proposed to utilize \textit{MicroNets} to construct asymmetric 3D ConvNets. 3D ConvNets benefit from end-to-end training, and requires only RGB input. Yet the temporal feature is extracted through pooling along the temporal dimension, and thus extracts only linear temporal feature. This causes part of the temporal feature might be lost during feature extraction operation.
\subsection{Correlation Modeling and Bilinear Pooling}\label{related:correlation}
Correlation modeling and bilinear pooling have been used in action recognition and have shown its success in improving temporal feature extraction. More specifically, Diba \textit{et al.} proposed TLE~\citep{diba2017deep} which represents the whole video by a bilinear model. The input of TLE is the temporal aggregated feature obtained by aggregating the features of each video segment. Meanwhile, Zhao \textit{et al.}~\citep{zhao2018recognize} utilizes correlation modeling for the construction of cost volume, which is an intermediate facility of optical flow estimation. 
More recently, inspired by the non-local mean operation for image denoising~\citep{buades2005non, li2016novel}, Wang \textit{et al.}~\citep{wang2018non} presented non-local operations to capture correlation on a pixel level as the representation of the temporal feature. Unlike the mentioned works above, our work utilizes correlation modeling on a frame-wise regional level, and computes the inter-frame correlation within a certain region through bilinear operation. This guarantees our temporal feature extraction method to be more computation efficient while maintaining improvement in temporal feature extraction.

\section{Method}
\label{section:method}

\begin{figure*}[t]
\begin{center}
    \includegraphics[width=1.\linewidth]{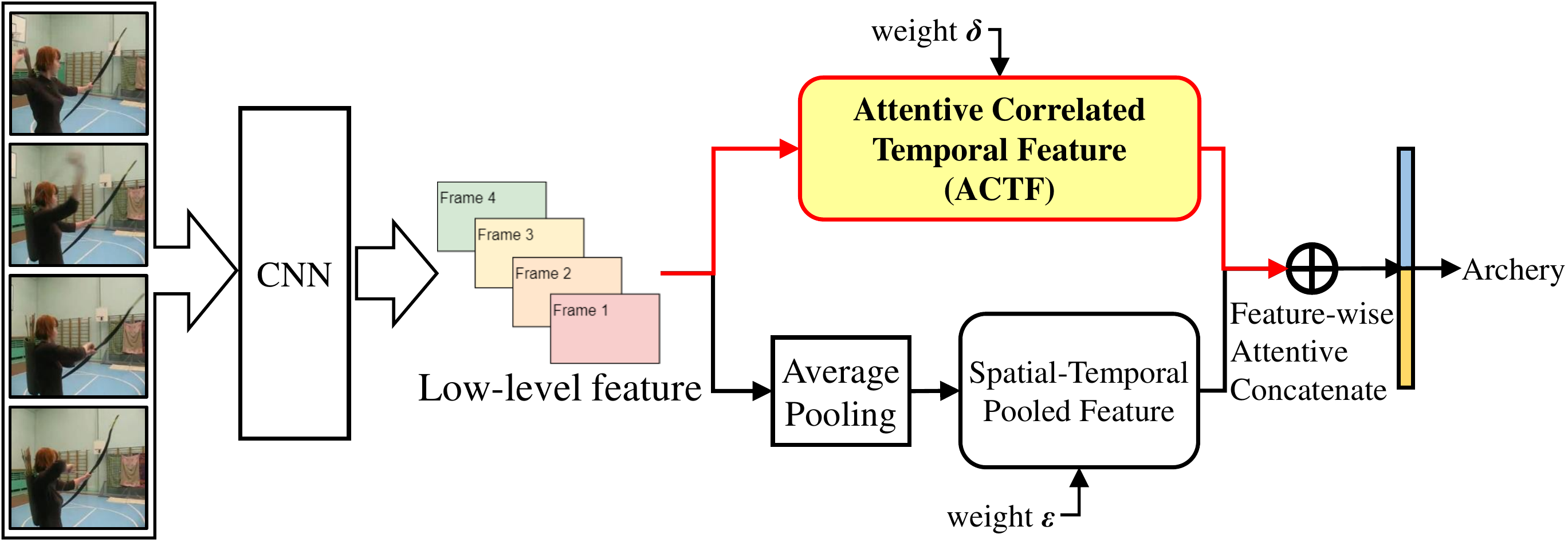}
\end{center}
   \caption{Detailed illustration of applying ACTF for action recognition. The sharp rectangles represent the networks or operations performed, while the rounded rectangles represent the resulting features. The overall framework takes the raw RGB frames as input. The low-level feature of the RGB frames is extracted through a ConvNet (CNN). From the low-level feature, we obtain the spatial-temporal pooled feature of the video through average pooling across both spatial and temporal dimensions. This feature is regarded as the spatial feature of the video. Simultaneously, we obtain the ACTF as the temporal feature of the video. Both features are combined attentively to form the whole representation of the video.}
\label{structure:overall}
\end{figure*}

The primary goal of our work is to develop an effective video-based action recognition framework with focus on temporal feature extraction. The main idea of the proposed method is to explore correlation of successive frames within a certain region, which captures temporal information. The extracted correlation feature can work with various low-level feature extraction networks that are normally convolutional neural networks (ConvNets), e.g. C3D and 3D-ResNet. These networks normally adopt a simple temporal pooling operation for obtaining the video representation. We propose an ACTF model to effectively extract the inter-frame corresponding-regional correlation feature and combine it with the feature obtained from simple temporal pooling. Next, we present a general action recognition framework that uses the proposed ACTF for temporal feature extraction, and then describe the details of the ACTF. The attention mechanism employed in ACTF will also be briefly explained.

\subsection{General Framework for Action Recognition with ACTF}
\label{framework:general}

The prominent methods for action recognition employ multiple modality networks, e.g. two-stream convolutional networks~\citep{NIPS2014_5353}. In these networks, temporal and spatial features are extracted and processed separately. Figure~\ref{structure:overall} shows the overall framework in our study. Given an input video as a sequence of frames, the low-level feature of each frame is first extracted through a convolutional neural network (ConvNet). The resulted low-level feature is denoted by $\mathbf{F}\in\mathbb{R}^{t\times C_{out}\times H\times W}$, where $t$ denotes the number of frames, $C_{out}$ denotes the number of channels, and $H$, $W$ are the height and width. Subsequently, we obtain two features from this low-level feature, namely the Spatial-Temporal Pooled feature, and the ACTF feature. The Spatial-Temporal Pooled feature $\mathbf{V}_{stpooled}$ is obtained by performing spatial-temporal average pooling over the low-level feature. The ACTF feature $\mathbf{V}_{actf}$ is obtained through an attentive concatenation of features obtained by performing both bilinear and linear operations on successive frames. Each of the two features characterizes a different perspective of the video. Performing average pooling over the low-level feature results in a feature that provides a general appearance pattern of the video. Thus the Spatial-Temporal Pooled feature as shown in Figure~\ref{structure:overall} is referred to as the spatial feature of the video in this paper. Meanwhile, the ACTF feature captures the correlation pattern of successive frames within a certain region, and is referred to as the temporal feature here. Both features have a dimension of $C_{out}$, {\it i.e.}  $\mathbf{V}_{actf}\in\mathbb{R}^{C_{out}}$, and $\mathbf{V}_{stpooled}\in\mathbb{R}^{C_{out}}$.

In certain scenarios, the spatial feature of the video is sufficient to produce satisfactory action recognition result. This occurs when certain action types are associated with certain visual elements. Meanwhile, in other scenarios, temporal feature play a more vital role. This occurs when visual elements of the video may appear in different actions. Thus, we adopt a feature-wise attentive concatenation method to dynamically combine the spatial and temporal features. 

\subsection{Extraction of ACTF feature}
\label{framework:actf}

\begin{figure}[t]
\begin{center}
    \includegraphics[width=1.\linewidth]{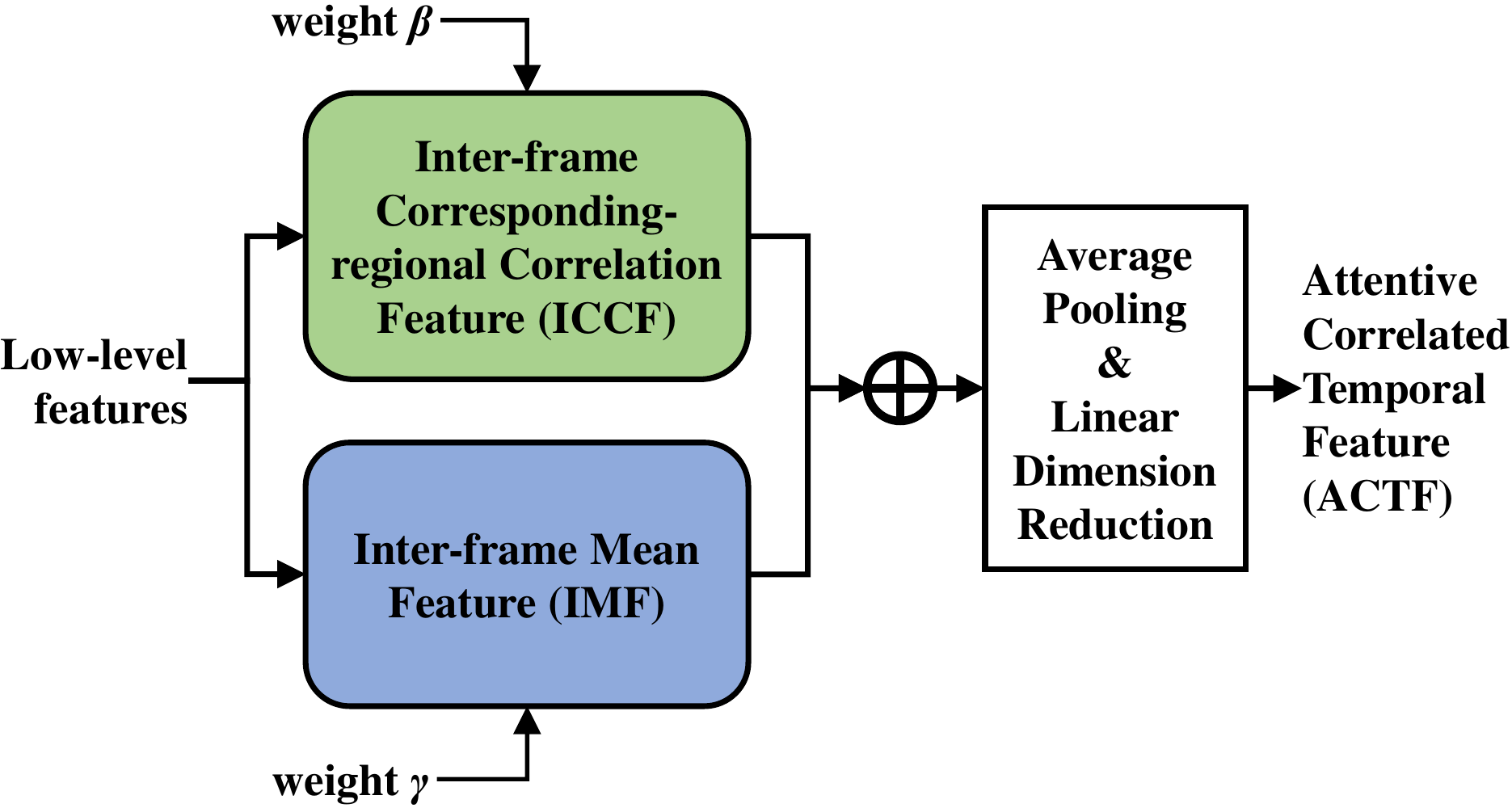}
\end{center}
   \caption{Illustration of the pipeline for extracting ACTF. From the low-level feature extracted, we extract two forms of inter-frame correlation features. A bilinear inter-frame correlation feature, extracted as the Inter-frame Corresponding-regional Correlation Feature (ICCF), as well as a linear inter-frame correlation feature, extracted as the Inter-frame Mean Feature (IMF). The features are combined attentively to form the ACTF.}
\label{structure:actf}
\end{figure}

Previous works~\citep{NIPS2014_5353,wang2016temporal,zhu2017hidden} show the importance of temporal feature in action recognition. However, most temporal information representations, such as optical flow used in two-stream convolutional network~\citep{NIPS2014_5353} or non-local operations in non-local 3D ConvNets~\citep{wang2018non}, are computationally expensive. This is due to the fact that both optical flow and non-local operations compute the correlations between successive frames on a pixel level. Computationally efficient RGB difference is employed in~\citep{wang2016temporal} to capture inter-frame relation, but shows inferior performance when used in combination with spatial feature. For slower actions where successive frames are very similar, RGB difference would return zero-valued correlation, and fail to capture the temporal feature, which might explain the inferior performance.

In this work, we propose to explore more sophisticated operations, such as bilinear function, on successive frames for temporal feature extraction. This is inspired by the bilinear operation used for fine-grained image recognition~\citep{7780410,lin2015bilinear}, where the within-image bilinear operation is used to learn local pairwise feature correlation through the outer product at every single position of the image. In this paper, the bilinear operation is extended across successive frames to discover inter-frame correlation within a certain region. Figure~\ref{structure:actf} shows the pipeline for extracting ACTF.

More specifically, given a video sequence, as described in Section~\ref{framework:general}, the low-level feature of the video is extracted through a ConvNet, whose output is $\mathbf{F}\in\mathbb{R}^{t\times C_{out}\times H\times W}$. We then extract a bilinear inter-frame correlation feature, the Inter-frame Corresponding-regional Correlation Feature (ICCF), and a linear inter-frame feature, the Inter-frame Mean Feature (IMF). The extraction function for the ICCF is denoted by $\mathcal{P}_{bilinear}$, while the extraction function for the IMF is denoted by $\mathcal{P}_{mean}$.

\begin{figure}[t]
\begin{center}
    \includegraphics[width=1.\linewidth]{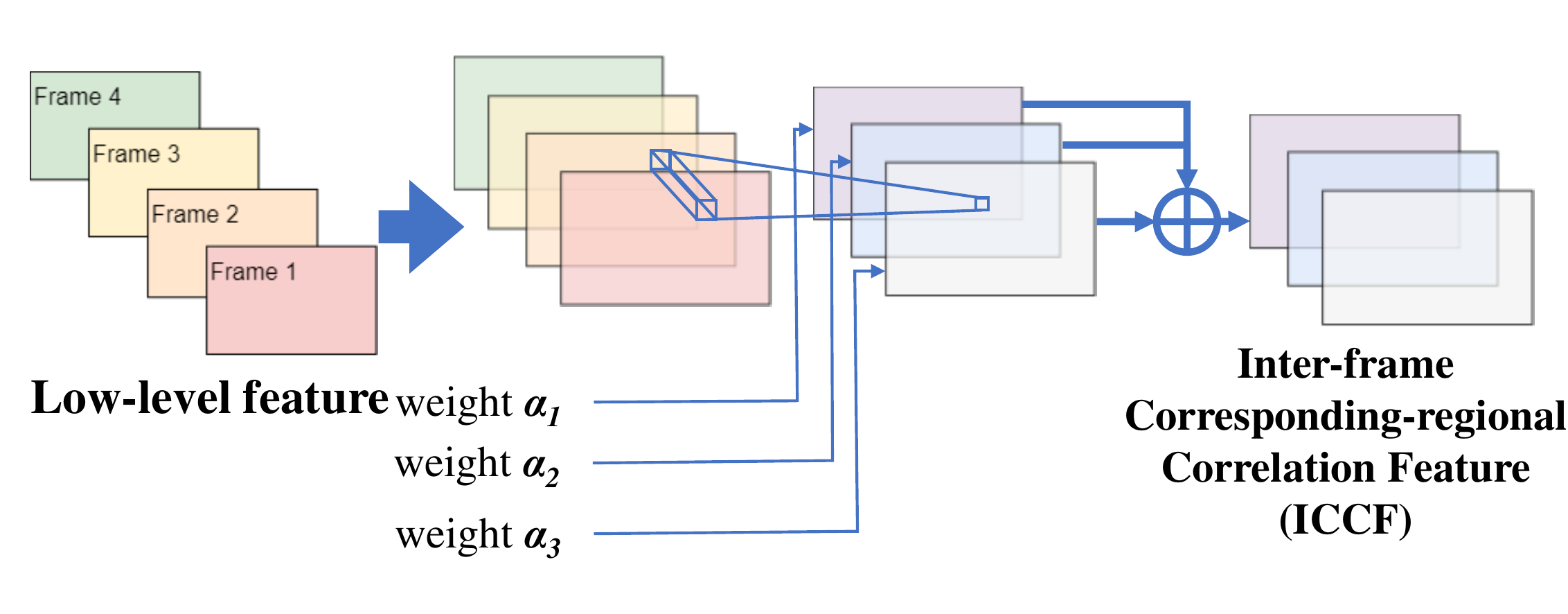}
\end{center}
   \caption{Illustration of the details for extracting the bilinear inter-frame correlation feature which is the ICCF. The pairwise bilinear correlation with respect to two successive frames within a certain region is computed for each pair of successive frames. The complete bilinear feature is extracted through temporal-wise attentive concatenation of each inter-frame correlation feature.}
\label{structure:iccf}
\end{figure}

We first describe $\mathcal{P}_{bilinear}$, which is the extraction function for the bilinear inter-frame correlation feature denoted as ICCF. Figure~\ref{structure:iccf} shows the details of extracting the ICCF. Denote $\mathbf{f}_{i}\in\mathbb{R}^{C_{out}\times H\times W}$ as the low-level feature extracted for frame $i$. To extract the bilinear inter-frame correlation feature, $\mathcal{P}_{bilinear}$ computes the pairwise bilinear correlation with respect to two successive frames within a certain region as follows:
\begin{equation}
\label{equation:bilinear}
\mathbf{b}_{i} = \mathcal{P}_{bilinear}(\mathbf{f}_{i}, \mathbf{f}_{i+1})
\end{equation}
Here $\mathbf{b}_{i}$ is the bilinear inter-frame correlation feature, and $\mathbf{b}_{i} \in\mathbb{R}^{C_{bilinear}\times H\times W}$, where $C_{bilinear}$  denotes the number of channels of the ICCF. 

More specifically, at the spatial location of $\mathcal{S}$, the feature of the current frame and the next frame is denoted as $\mathbf{f}_{i, \mathcal{S}}$ and $\mathbf{f}_{i+1, \mathcal{S}}$. We denote the bilinear operation function at location $\mathcal{S}$ to be $\mathcal{B}_{\mathcal{S}}$, and is formulated by the following equation:
\begin{equation}
\label{equation:bilinear-specific}
\mathcal{B}_{i, \mathcal{S}} = \mathbf{f}_{i, \mathcal{S}}{\mathbf{f}_{i+1, \mathcal{S}}}^{T}
\end{equation}
At the spatial location $\mathcal{S}$, the feature for frame $i$ is of size $C_{out}\times 1$. Thus, from Equation~\ref{equation:bilinear-specific}, the bilinear inter-frame correlation feature at location $\mathcal{S}$ is of size ${C_{out}}\times {C_{out}}$. We then reshape it such that the result would be of size ${C_{out}}^2\times 1$.

Although the bilinear inter-frame correlation feature obtained through Equation~\ref{equation:bilinear-specific} is direct, such feature representation is very high-dimensional. In our case where $C_{out}$ is around 750, the dimension of the bilinear inter-frame correlation feature at each spatial location is more than 500,000. Such high dimensional representation is impractical. Therefore, to obtain the desired bilinear correlation, we adopt a compact form of bilinear operation as implemented in~\citep{7780410}. 

The basis of the compact form of the bilinear operation is to find a low dimension projection function of $\mathcal{B}_{i, \mathcal{S}}$, denoted as $\mathcal{C}_{i, \mathcal{S}}$. The two functions are equivalent with respect to a linear kernel machine. Given two pairs of frames: frames $(i, i+1)$ and frames $(j, j+1)$, a linear kernel machine is formulated as:
\begin{equation}
    \begin{aligned}
    \label{equation:bilinear-equivalent}
    \langle\mathcal{B}_{i, \mathcal{S}}, \mathcal{B}_{j, \mathcal{S}}\rangle &= \langle\mathbf{f}_{i, \mathcal{S}}{\mathbf{f}_{i+1, \mathcal{S}}}^{T}, \mathbf{f}_{j, \mathcal{S}}{\mathbf{f}_{j+1, \mathcal{S}}}^{T}\rangle \\
    &=\langle\mathbf{f}_{i, \mathcal{S}}, \mathbf{f}_{j, \mathcal{S}}\rangle^2 
    \end{aligned}
\end{equation}
We then find a low dimension projection function as $\phi(\mathbf{f}_{i, \mathcal{S}})\in\mathbb{R}^{d}$ such that $\langle\phi(\mathbf{f}_{i, \mathcal{S}}), \phi(\mathbf{f}_{j, \mathcal{S}})\rangle \approx k(\mathbf{f}_{i, \mathcal{S}}, \mathbf{f}_{j, \mathcal{S}})$, where $k$ is a polynomial kernel. Such projection function $\phi(\mathbf{f}_{i, \mathcal{S}})$ would allow us to approximate Equation~\ref{equation:bilinear-equivalent} by:
\begin{equation}
    \begin{aligned}
    \label{equation:bilinear-compact}
    \langle\mathcal{B}_{i, \mathcal{S}}, \mathcal{B}_{j, \mathcal{S}}\rangle &=\langle\mathbf{f}_{i, \mathcal{S}}, \mathbf{f}_{j, \mathcal{S}}\rangle^2 \\
    &\approx\langle\phi(\mathbf{f}_{i, \mathcal{S}}), \phi(\mathbf{f}_{j, \mathcal{S}})\rangle \\
    &\equiv\langle\mathcal{C}_{i, \mathcal{S}}, \mathcal{C}_{j, \mathcal{S}}\rangle
    \end{aligned}
\end{equation}
where $\mathcal{C}_{i, \mathcal{S}} = \phi(\mathbf{f}_{i, \mathcal{S}})$ is the compact form of the bilinear operation $\mathcal{B}_{i, \mathcal{S}}$. Hence to obtain the compact form, we need to find the low dimension approximation of the polynomial kernel $k$. Here we utilize the Tensor Sketch approximation method proposed in~\citep{Pham:2013:FSP:2487575.2487591}. Ultimately, our extraction function $\mathcal{P}_{bilinear}$ computed at each spatial location $\mathcal{S}$ for frame $i$ and the successive frame $i+1$ is equivalent to its compact form $\mathcal{C}_{i, \mathcal{S}}$.

For image recognition tasks, features extracted through bilinear function go through a sum pooling operation to extract the complete representation of the image. However, if such a pooling method is used in videos, the temporal information may be lost. This conflicts with our goal of extracting temporal information through the bilinear inter-frame correlation feature. To dynamically combine all bilinear inter-frame correlation features temporally, we apply a temporal-wise attentive concatenation to each pair of successive frames. A learnable weight parameter $\alpha_{i}$ is assigned to each inter-frame correlation feature $\mathbf{b}_{i}$. Such attentive concatenation allows the extracted ICCF to focus on the pair of frames where the action most likely takes place. The result is a feature $\mathbf{B}\in\mathbb{R}^{(t-1)\times C_{bilinear}\times H\times W}$. For each pair of successive frames, $\mathbf{B}_i=\alpha_{i}\mathbf{b}_i$.

To extract the temporal feature of the video more accurately, besides the bilinear inter-frame correlation feature, we would also need the linear inter-frame correlation feature denoted as IMF. The IMF provides a baseline for the bilinear inter-frame correlation feature, and is important when actions are similar temporally but very different in appearance. Following this idea, we feed the low-level feature $\mathbf{F}$ to extract the IMF in parallel with the ICCF. 
Unlike common temporal pooling layers where the average pooling is performed across the whole temporal dimension, the purpose of the extraction function $\mathcal{P}_{mean}$ is to capture the average of two successive frames. More specifically, the extraction function for IMF is an average pooling function with a kernel size of $(k_i, k_h, k_w)$. Here $k_h$ and $k_w$ are the kernel size corresponding to the spatial dimensions. As we need to preserve all information along the spatial dimensions, hence $k_h, k_w=1$. To obtain the average pooling along successive temporal features $\mathbf{f}_i$ and $\mathbf{f}_{i+1}$, the kernel size along the temporal dimension $k_i$ is set to 2 instead of the whole temporal dimension length.
The IMF $\mathbf{L}\in\mathbb{R}^{(t-1)\times C_{out}\times H\times W}$ is computed by:
\begin{equation}
\label{equation:inter-average}
\mathbf{L} = \mathcal{P}_{mean}(\mathbf{F})
\end{equation}

Similar to the temporal-wise attentive concatenation for bilinear inter-frame correlation features, we adopt a feature-wise attentive concatenation approach to combine the bilinear inter-frame correlation feature with the linear inter-frame correlation feature. Each of the two types of features is assigned a separate weight parameter, denoted as $\beta, \gamma$ respectively. This allows the network to dynamically focus on either feature for different actions. The result of the attentive concatenation $\mathbf{H}\in\mathbb{R}^{(t-1)\times C_{concat}\times H\times W}$ is obtained as follows:
\begin{equation}
\label{equation:concat-actf}
\mathbf{H} = \beta\mathbf{B}\oplus\gamma\mathbf{L}
\end{equation}
where $\oplus$ denotes the concatenation operation along the feature channel dimension. $C_{concat}$ is the total number of feature channels, which is the sum of $C_{out}$ and $C_{bilinear}$.

The complete ACTF feature $\mathbf{V}_{actf}$ is obtained as follows:
\begin{equation}
\label{equation:final-rep}
\mathbf{V}_{actf} = \mathcal{L}(\mathcal{P}_{average}(\mathbf{H}))
\end{equation}
Where $\mathcal{P}_{average}$ is an average pooling operation with a kernel size of $((t-1), H, W)$ corresponding to the temporal and spatial dimensions respectively. ACTF feature summarizes both the ICCF and the IMF. $\mathcal{L}$ is a linear dimension reduction function, constructed as a multi-layer linear neural network. This allows $\mathcal{L}$ to be learnable and the overall system to be trainable in an end-to-end manner. The resulting feature is thus the ACTF feature $\mathbf{V}_{actf} \in\mathbb{R}^{C_{out}}$.

\subsection{Attentive Concatenation of Features}
\label{framework:attentive}

Our network is designed to focus on the pairs of time steps which are more relevant to the action. Meanwhile, it is also designed to focus on the more important type of feature, {\it i.e.} the spatial or temporal feature . To achieve both goals, we adopt attentive concatenation at each location where different features are combined. In this section, we describe how to extract the ICCF by the temporal-wise attentive concatenation of bilinear inter-frame correlation features. The combination of features $\mathbf{V}_{stpooled}$ and $\mathbf{V}_{actf}$ mentioned in Section~\ref{framework:general} as well as the combination of features $\mathbf{B}$ and $\mathbf{L}$ mentioned in Section~\ref{framework:actf} follow similar implementations.

The attentive concatenation of all $(t-1)$ bilinear correlation features is achieved by assigning each feature with a weight $\alpha_{i}$ for the $i^{th}$ correlation feature $\mathbf{b}_{i}$. Inspired by the cascade attention network proposed in~\citep{Wang:2018:CAN:3242969.3264991}, we adopt an attentive concatenation approach for the computation of weight $\alpha_i$. Formally, $\alpha_{i}$ is computed by:
\begin{equation}
\label{equation:weight-alpha}
\alpha_{i} = g(h((\mathcal{P}_{spatial}(\mathbf{b}_{i}))W))
\end{equation}
More specifically, given the $i^{th}$ bilinear correlation feature $\mathbf{b}_i \in\mathbb{R}^{C_{bilinear}\times H\times W}$, $\mathcal{P}_{spatial}$ is a spatial average pooling function with kernel size $(H, W)$. The output of $\mathcal{P}_{spatial}$ is a pooled feature vector $\mathbf{b}_{pooled,i}\in\mathbb{R}^{C_{bilinear}}$. $W\in\mathbb{R}^{C_{bilinear}\times 1}$ denotes a trainable parameter matrix, shared among all $(t-1)$ bilinear correlation features. The result of this matrix multiplication is a primitive weight parameter denoted as $\alpha_{prime, i}$.

To scale the primitive weight parameter $\alpha_{prime, i}$ to a range of $[0, 1]$, we apply a sigmoid function denoted as $h(\alpha_{prime, i})$, which is computed by:
\begin{equation}
\label{equation:sigmoid}
h(\alpha_{prime, i}) = \frac{\mathrm{1}}{\mathrm{1}+e^{-{\alpha_{prime, i}}}}
\end{equation}
The weight $\alpha_{i}$ is then further processed from $h(\alpha_{prime, i})$ to satisfy $\Sigma\alpha_{i}=1$. This is achieved by applying a softmax function denoted by $g(\cdot)$, and the weight $\alpha_{i}$ is calculated as follows:
\begin{equation}
\label{equation:softmax}
    \begin{aligned}
    \alpha_{i} &= g(h(\alpha_{prime, i})) \\
    &= \frac{e^{h(\alpha_{prime, i})}}{\sum\nolimits_{i=1}^{t-1}{e^{h(\alpha_{prime, i})}} }
    \end{aligned}
\end{equation}
The weight $\alpha_i$ indicates the importance of the $i^{th}$ bilinear inter-frame correlation feature, $\mathbf{b}_{i}$.

\section{Experiments}
\label{section:experiment}

In this section, we present our evaluation results of the proposed work. The evaluation is conducted through action recognition experiments on two public benchmark datasets. We present state-of-the-art results on a competitive architecture, and prove the novelties on another similar baseline. We also present detailed ablation study of the components of our proposed framework to verify our design.

\subsection{Experimental Settings}
\label{experiment:settings}

We conduct experiments on two benchmark datasets of action recognition: UCF101~\citep{soomro2012ucf101} and HMDB51~\citep{kuehne2011hmdb}. The UCF101 dataset contains 13,320 videos from 101 action categories while the HMDB51 dataset contains 6,766 videos from 51 action categories. We follow the experiment settings as in~\citep{10.1007/978-3-030-01246-5_22,Tran:2015:LSF:2919332.2919929,tran2018closer} that adopt the three training/testing splits for evaluation. We report the average top-1 accuracy of the three splits. Our proposed framework for temporal feature extraction can be used in any ConvNet based networks. To obtain the state-of-the-art result, we instantiate MFNet~\citep{10.1007/978-3-030-01246-5_22}. 

Our experiments are implemented using PyTorch~\citep{paszke2017automatic}. Following the implementation in~\citep{10.1007/978-3-030-01246-5_22}, the input is a frame sequence with each frame of size $224\times224$. The output from MFNet~\citep{10.1007/978-3-030-01246-5_22} is a low-level feature of size $8\times768\times 7\times 7$, where the number of output channels is $768$ . Each frame is represented by a feature of size $7\times7$. We set the number of channels of our ICCF to 3,840. Thus, the size of $\mathbf{H}$, described in Section 3.2 of the paper, is $7\times4608\times7\times7$. We design the linear dimension reduction function in Equation 7 as a three-layer linear neural network with RELU activation. For training, we utilize the pretrained model of MFNet~\citep{10.1007/978-3-030-01246-5_22} trained on Kinetics~\citep{kay2017kinetics}, a large-scale human action dataset. To accelerate our training, the pretrained model is used for the initialization of the network which includes our framework for temporal feature extraction. We use stochastic gradient descent algorithm~\citep{bottou2010large} for optimization, setting the weight decay to 0.0001 and the momentum to 0.9. For both datasets, our initial learning rate is set to 0.005. For UCF101~\citep{soomro2012ucf101} dataset, the learning rate is decreased for four times, while for HMDB51~\citep{kuehne2011hmdb} dataset, the learning rate is decreased for three times. The learning rate is decreased with a factor of 0.1.

To prove that our approach can be applied to other 3D ConvNet approaches, we also apply our proposed ACTF in another 3D ConvNet. We instantiate C3D~\citep{Tran:2015:LSF:2919332.2919929}, a classical 3D ConvNet baseline for action recognition. Our proposed ACTF is extracted after conv5 layer of the C3D network, in parallel with the spatial-temporal pooling layer pool5, as well as the linear layer that follows. We follow the setup as in~\citep{Tran:2015:LSF:2919332.2919929}, using stochastic gradient descent~\citep{bottou2010large} with initial learning rate of 0.001. We compare the results of the C3D network with and without the temporal feature extracted by our proposed framework on HMDB51 dataset.

\subsection{Results and Comparison}
\label{experiment:results-comparion}
Table~\ref{table:compare-sota} shows the comparison of top-1 accuracy on UCF101 and HMDB51 datasets with other state-of-the-art methods including:

\begin{enumerate}
        \item \textbf{Two-stream methods: }the original two-stream method (original TS)~\citep{NIPS2014_5353}, Hidden Two-Stream (Hidden TS)~\citep{zhu2017hidden}, Long-term Temporal Convolutions (LTC)~\citep{varol18_ltc}, ActionVLAD~\citep{girdhar2017actionvlad} and Temporal Segment Network (TSN)~\citep{wang2016temporal}
        \item \textbf{3D ConvNets-based methods: }C3D~\citep{Tran:2015:LSF:2919332.2919929}, TSN with RGB input~\citep{wang2016temporal}, Res3D~\citep{tran2017convnet}, ST-ResNet~\citep{NIPS2016_6433}, 3D-ResNext~\citep{hara2018can}, R(2+1)D with RGB input~\citep{tran2018closer}, I3D with RGB input~\citep{carreira2017quo}, TVNet~\citep{fan2018end}, MFNet~\citep{10.1007/978-3-030-01246-5_22} and T-C3D~\citep{liu2018t}
\end{enumerate}

Our state-of-the-art performance is achieved by instantiating MFNet, denoted as MFNet-ACTF. For this experiment, we set our batch size to 80 and conduct the experiment using four NVIDIA Tesla P100 GPUs. 

\begin{table}
\small
\centering
\smallskip\begin{tabular}{c|c|c|c|c}
\hline
\hline
& Method & UCF101 & HMDB51 & FPS\\
\hline
\multirow{4}{*}{Two-stream}& original TS & 88.0\% & 59.4\% & 14\\
& Hidden TS & 90.3\% & 58.9\% & $<\!14$\\
& LTC & 91.7\% & 64.8\% & $<\!14$\\
& TSN & 94.2\% & 69.4\% & 5\\
\hline
\multirow{10}{*}{3D ConvNets}&C3D & 85.2\% & 65.5\% & 314\\
& TSN (RGB) & 86.2\% & - & N/A\\
& Res3D & 85.8\% & 54.9\% & N/A\\
& T-C3D & 91.8\% & 62.8\% & 969\\
& ST-ResNet & 93.5\% & 66.4\% & N/A\\
& 3D-ResNext & 94.5\% & 70.2\% & $<\!314$\\
& R(2+1)D (RGB) & 93.6\% & 66.6\% & N/A\\
& I3D (RGB) & 95.6\% & 74.8\% & N/A\\
& TVNet & 95.4\% & 72.5\% & N/A\\
& MFNet & 96.0\% & 74.6\% & N/A\\
\hline
Ours & \textbf{MFNet-ACTF}& \textbf{96.3\%} & \textbf{76.3\%} & \textbf{478}\\
\hline
\hline
\end{tabular}
\caption{Comparison of top-1 accuracy and speed with state-of-the-art methods on UCF101 and HMDB51 datasets.}
\label{table:compare-sota}
\end{table}

The performance results in Table~\ref{table:compare-sota} show that our network achieves the best results on both benchmark datasets. More specifically, our MFNet-ACTF network achieves a $1.7\%$ improvement on HMDB51 dataset over the networks whose input are solely RGB frames. Our method even surpasses several networks with both RGB and optical flow as input. For UCF101 dataset, our MFNet-ACTF also produces the best result. It is noted that the improvement is not as significant as that on HMDB51 dataset, mainly due to the fact that there is little room for improvement.

The speed results in Table~\ref{table:compare-sota} show that our proposed method balances between high accuracy and relatively high inference speed. Despite achieving high accuracy on both dataset, two-stream methods such as TSN could not achieve real-time requirements, reaching only 5 FPS. Compared with two-stream methods, our proposed method is much faster in inference speed, reaching a speed of 478 FPS, which is well above real-time requirements. Our speed is even faster than that achieved by C3D network. Note that our speed is slower than that achieved by T-C3D network, but we achieved a much higher accuracy compared to theirs, with a $13.5\%$ increase in top-1 accuracy on HMDB51 dataset.

We suggest in Section~\ref{section:method} that our proposed framework which includes ACTF feature can be used in combination with any ConvNet-based low-level feature extraction networks, such as C3D network. To verify this, we conducted experiments on the baseline network C3D with and without ACTF. We first perform action recognition with only the temporal feature extracted through our proposed ACTF. The low-level feature is extracted through conv5 layer of the C3D network. We denote this network as C3D-single-ACTF. We then perform action recognition by attentively combining the ACTF feature as well as the Spatial-Temporal Pooled feature which is extracted from \textit{pool5} layer of C3D, similar to the implementation of MFNet-ACTF. We denote this modification as C3D-ACTF. The top-1 accuracy of the networks are shown in Table~\ref{table:compare-c3d}.

\begin{table}
\centering
\smallskip\begin{tabular}{l|c}
\hline
\hline
Method & Top-1 HMDB51\\
\hline
C3D & 65.5\%\\
\hline
C3D-single-ACTF & 67.9\%\\
C3D-ACTF & 69.2\%\\
\hline
\hline
\end{tabular}
\caption{Top-1 accuracy of C3D network on HMDB51 dataset with and without our proposed framework.}
\label{table:compare-c3d}
\end{table}

The results in Table~\ref{table:compare-c3d} clearly show that applying our proposed framework in the C3D network improves the accuracy of the baseline C3D network. Even by only utilizing the extracted ACTF feature as temporal feature, we obtain an improvement of $2.4\%$. This shows that our ACTF feature effectively represent the temporal pattern in the video and thus lead to better results. A larger gain of $3.7\%$ is achieved when the extracted ACTF feature is used with the Spatial-Temporal Pooled feature. The results are consistent with that shown in Table~\ref{table:compare-sota}, where using ACTF feature improves the accuracy of MFNet. This suggests that our proposed framework is generic, and can be used with other baselines.

\begin{figure}[t]
\begin{center}
    \includegraphics[width=1.\linewidth]{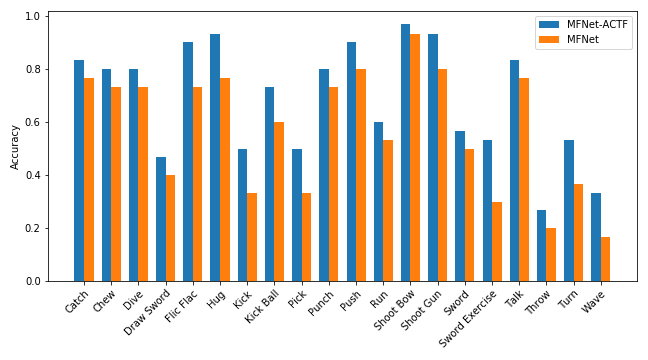}
\end{center}
   \caption{Accuracy comparisons of 20 classes on split 1 of the HMDB-51 between our proposed MFNet-ACTF network and the original MFNet network.}
\label{result:per-class-comparison}
\end{figure}

\begin{figure}[t]
\begin{center}
    \includegraphics[width=1.\linewidth]{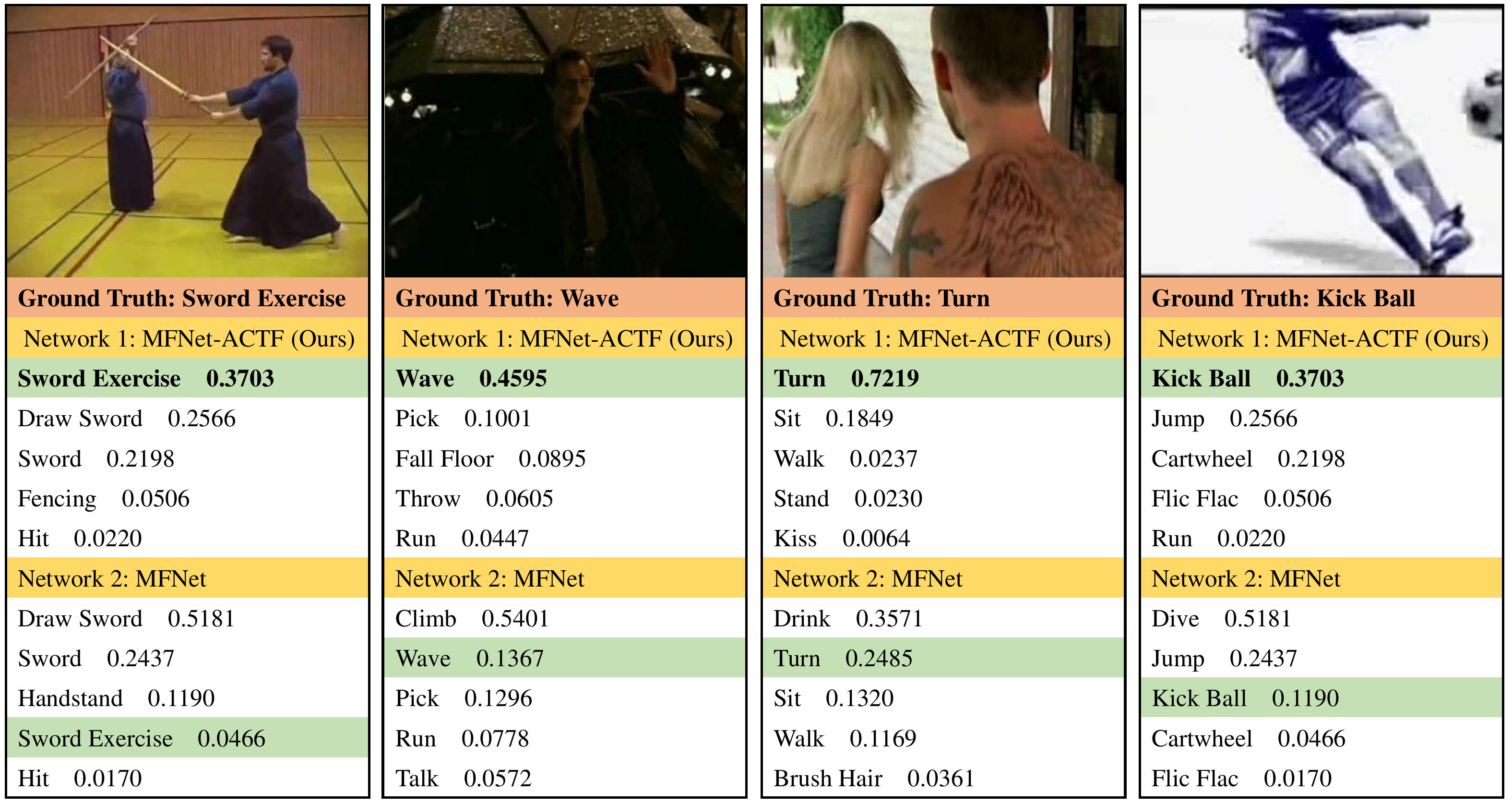}
\end{center}
   \caption{Examples from HMDB51 dataset where our proposed MFNet-ACTF succeeds in recognizing the action while the original MFNet fails.}
\label{result:examples-comparison}
\end{figure}
 
We further investigate the improvement of performance over different actions and present the comparison of performance between our proposed MFNet-ACTF network and the original MFNet network. Figure~\ref{result:per-class-comparison} shows the accuracy of 20 classes from split-1 of the HMDB51 dataset, where our network outperforms the original network by a noticeable margin. It is worth noticing that for actions with similar spatial appearance but different actions, e.g. "Sword" and "Sword Exercise", our network performs significantly better than the original network. Our network obtains a $23.3\%$ performance gain over the original MFNet on the action class "Sword Exercise". The large performance gain proves the effectiveness of the additional temporal feature extracted as the ACTF feature in improving the complete video representation. 
Several examples from HMDB51 dataset is presented in Figure\ref{result:examples-comparison} where our proposed MFNet-ACTF could accurately recognize the respective actions while the original MFNet network could not. It could be observed that the spatial features of the given examples, or more intuitively the appearance of the given examples, could not provide effective representation for accurate action recognition. For example, for the first video, the scenario as shown could be present in action classes "Sword", in which most videos present people fighting with a sword, and "Draw Sword", in which videos present the action of a sword drawn out. The difference between these action classes could only be determined through the temporal feature instead of the spatial feature. Thus the original network which can only extract the spatial feature of the video cannot distinguish the actions correctly while our proposed framework succeeds in recognizing the different actions.
\subsection{Ablation Study}
\label{experiment:ablation}

\begin{table}
\centering
\begin{tabular}{l|c}
\hline
\hline
Method & Top-1 HMDB51 split-1\\
\hline
MFNet & 70.8\%\\
\hline
MFNet-ICCF & 72.6\%\\
MFNet-single-ACTF & 72.9\%\\
\hline
\hline
\end{tabular}
\caption{Comparison of the network architectures that use only temporal feature for action recognition.}
\label{table:compare-iccf-actf}
\end{table}

\begin{table}
\centering
\begin{tabular}{l|c}
\hline
\hline
Method & Top-1 HMDB51 split-1\\
\hline
MFNet-ACTF-no-attn & 72.5\%\\
MFNet-attn@ACTF & 73.3\%\\
MFNet-attn@final & 73.0\%\\
\hline
MFNet-ACTF & 73.6\%\\
\hline
\hline
\end{tabular}
\caption{Comparison of the network architectures that use all or partial attentive concatenation.}
\label{table:compare-attentive}
\end{table}

In this section, we justify our proposed design of the ACTF feature through ablation study. Specifically, we examine the performance of our proposed ICCF and the ACTF feature separately. We then examine the performance of the attention mechanisms used to combine the different modules of our proposed generic action recognition framework as discussed in Section~\ref{section:method}. All experiments conducted in our ablation study are performed on split 1 of the HMDB51 dataset. We set our batch size to 16 and conduct the experiment using one NVIDIA TITAN Xp GPU. The much smaller batch size is a key reason of the lower accuracy reported than that in Table~\ref{table:compare-sota}.

We instantiate MFNet to justify our proposed ICCF and the ACTF feature introduced in Section~\ref{framework:actf} and utilize only temporal feature for action recognition. First the proposed ICCF is extracted as our temporal feature. The network that utilizes only ICCF is denoted as MFNet-ICCF. We then employ the ACTF feature as our temporal feature. Similar to the previous denotation, the network that utilizes only the ACTF feature is denoted as MFNet-single-ACTF. The comparison of the performances of these two networks with the baseline MFNet is shown in Table~\ref{table:compare-iccf-actf}.

The result in Table~\ref{table:compare-iccf-actf} shows that by utilizing only temporal feature, even with only bilinear inter-frame correlation, the performance of the network is improved by a margin of $1.8\%$, indicating that utilizing inter-frame correlation information helps to extract high-quality temporal feature of the video. The improvement achieved by using high quality temporal feature over feature obtained from spatial-temporal pooling coincides with findings in preceding works~\citep{wang2016temporal,carreira2017quo,7780582}. However, our temporal feature is obtained from RGB input through inter-frame correlation rather than using optical flow.

Table~\ref{table:compare-iccf-actf} shows further improvement when we employ the ACTF feature. As described in Section~\ref{framework:actf}, the ACTF feature is a weighted combination of ICCF, which is a bilinear inter-frame correlation feature, and IMF, which is a linear inter-frame correlation feature. This result proves that the bilinear inter-frame correlation feature and linear inter-frame feature complements each other.

To better combine the features extracted from different modules, we introduced attentive concatenation of features as mentioned in Section~\ref{framework:attentive}. Here we justify the need for utilizing attentive concatenation of features. Table~\ref{table:compare-attentive} presents the comparison between the networks that utilize attentive concatenation at every step and the networks that partially or do not utilize attentive concatenation for feature combination. Here MFNet-ACTF-no-attn denotes the network where all feature combination utilizes direct concatenation instead of attentive concatenation. Meanwhile, MFNet-ACTF denotes the network utilizing our proposed temporal feature extraction framework with attentive concatenation at every step of feature combination. MFNet-attn@ACTF denotes the network that performs temporal-wise attentive concatenation when constructing ICCF, and feature-wise attentive concatenation of ICCF and IMF as shown in Figure~\ref{structure:actf}. The concatenation of the temporal feature and spatial feature is by direct concatenation. Similarly, MFNet-attn@final denotes that  attentive concatenation is adopted only for ACTF and Spatial-Temporal Pooled feature combination while direct concatenation is adopted at other stages.

The result given in Table~\ref{table:compare-attentive} clearly shows the advantage of adopting attentive concatenation for feature combination. We note that if the network combines features with only direct concatenation, its performance would be even worse than that of MFNet-single-ACTF, whose ACTF feature is constructed with attentive concatenation of ICCF and IMF features. The performance is improved even when attentive concatenation is used in some stages of feature combination only. It can be observed that applying attentive concatenation at different stages complements each other, with over $1\%$ improvement made when all stages adopt attentive concatenation. 

\begin{figure}[t]
\centering
    \includegraphics[width=1.\linewidth]{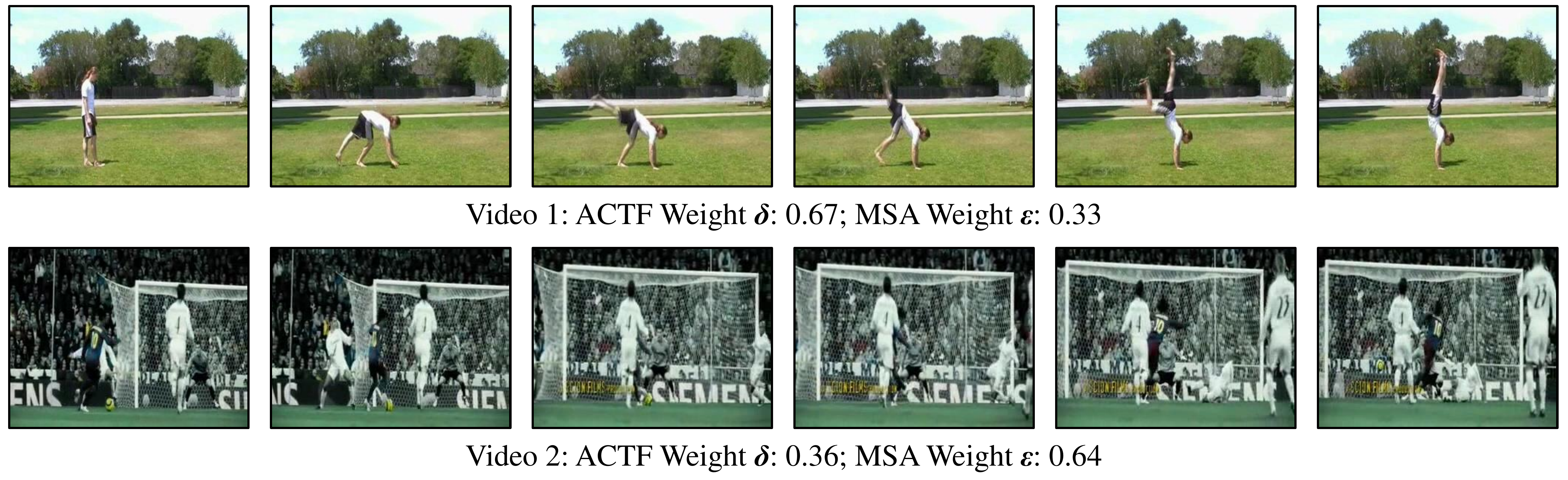}
   \caption{The weights of ACTF feature $\delta$ and the weights of Spatial-Temporal Pooled feature $\epsilon$ for two videos. Attentive concatenation learns these weights dynamically.}
\label{result:delta-epsilon}
\end{figure}

We also investigate the weights $\delta$ and $\epsilon$ on the ACTF feature and the Spatial-Temporal Pooled feature for different videos. Figure~\ref{result:delta-epsilon} shows examples where either temporal feature or spatial feature dominates the feature combination process. Video 2 shows a video where the Spatial-Temporal Pooled feature, or the spatial feature, dominates the feature combination. These videos tend to have clear visual characteristics, such as the soccer goal that appears in most videos describing the sport soccer. The appearance of these videos are therefore sufficient for action recognition, and dominate the feature combination process. By contrast, feature combination in Video 1 is dominated by ACTF feature, which is the temporal feature. We observe that similar videos tend to have actions that would mix up with other categories. In this case, the handstand action is similar to actions that may occur in diving or in somersault, where a person would also go upside down. Also, there is no iconic background items in Video 1. For these videos, the temporal features dominate the feature combination, thus has a larger weight $\delta$. The different weights with respect to the different videos could prove that adopting attentive concatenation could attend to the more important feature which is related to the characteristic of the video itself.

\section{Conclusion}
\label{section:conclusion}

In this work, we propose a new method for extracting the temporal feature of a video while avoiding the use of optical flow. The new temporal feature namely Attentive Correlated Temporal Feature (ACTF) is an attentive combination of both bilinear inter-frame correlation and linear inter-frame correlation features. The bilinear inter-frame correlation feature is extracted through a bilinear operation with respect to successive frames within a certain region, while the linear inter-frame feature is extracted through inter-frame temporal pooling. For overall evaluation on UCF101 and HMDB51, our method obtains state-of-the-art results when instantiating MFNet combined with our ACTF feature. We verify our design through thorough ablation study, and then further demonstrate that the proposed feature can be introduced to other similar action recognition networks instead of using optical flow. 

\section*{References}
\bibstyle{elsarticle-num}
\bibliography{egbib}

\end{document}